\documentclass[12pt]{article}
\usepackage[margin=1in]{geometry}
\usepackage{graphicx}
\usepackage{tikz}
\usepackage{standalone}
\usetikzlibrary{positioning,fit,calc,arrows.meta}
\usepackage{amsmath,amssymb,bm}
\usepackage{booktabs}
\usepackage[table]{xcolor}
\usepackage{tabularx}
\usepackage{setspace}
\usepackage{hyperref} 
\usepackage{standalone}
\usepackage[backend=biber,style=numeric,autocite=superscript]{biblatex}
\title{Generalized Machine Learning for Fast Calibration of Agent-Based Epidemic Models\\
}
\author{Sima Najafzadehkhoei \and George G. Vega Yon \and Derek S. Meyer \and Bernardo Modenesi}
\date{March 13, 2026}

\begin{document}
\maketitle

\begin{abstract}
Agent-based models (ABMs) are widely used to study infectious disease dynamics, but their calibration is often computationally intensive, limiting their applicability in time-sensitive public health settings. We propose DeepIMC (Deep Inverse Mapping Calibration), a machine learning–based calibration framework that directly learns the inverse mapping from epidemic time series to epidemiological parameters. DeepIMC trains a bidirectional Long Short-Term Memory (BiLSTM) neural network on synthetic epidemic trajectories generated from agent-based models such as the Susceptible–Infected–Recovered (SIR) model, enabling rapid parameter estimation without repeated simulation at inference time.
We evaluate DeepIMC through an extensive simulation study comprising 5,000 heterogeneous epidemic scenarios and benchmark its performance against Approximate Bayesian Computation (ABC) using likelihood-free Markov Chain Monte Carlo. The results show that DeepIMC substantially improves parameter recovery accuracy, produces sharp and well-calibrated predictive intervals, and reduces computational time by more than an order of magnitude relative to ABC. Although structural parameter identifiability constraints limit the precise recovery of all model parameters simultaneously, the calibrated models reliably reproduce epidemic trajectories and support accurate forward prediction with their estimated parameters.
DeepIMC is implemented in the open-source R package \texttt{epiworldRCalibrate}, facilitating practical adoption for real-time epidemic modeling and policy analysis. Overall, our findings demonstrate that DeepIMC provides a scalable, operationally effective alternative to traditional simulation-based calibration methods for agent-based epidemic models.
\end{abstract}
\section{Introduction}
Agent-based models (ABMs) are increasingly valuable tools for understanding complex systems, particularly in public health contexts where simulating interactions among individuals can reveal important epidemiological dynamics. However, calibrating ABMs—identifying parameter values that produce realistic model outputs—is typically computationally intensive, posing a significant challenge for timely public health responses. This paper introduces a machine learning (ML) algorithm designed to address these calibration challenges effectively. By leveraging training data generated directly from the ABM and performing supervised learning with it, our approach not only improves calibration accuracy but also significantly reduces the computational burden. In outbreak and forecasting settings, where decisions must be updated rapidly as new data arrive, the computational burden of traditional calibration methods can severely limit the practical use of agent-based models.

Although various methods integrate ML with ABMs \autocite{Angione2022-surrogate} \autocite{lamperti2017agentbasedmodelcalibrationusing}, most, if not all, are utilized to create mappings from observed parameters to realized data, often referred to as surrogate models. In contrast, our method generates the inverse mapping: from observed data back to the parameters. From an epidemiological perspective, this inverse mapping enables rapid recovery of dynamically equivalent parameter sets that reproduce observed incidence trajectories, which is often sufficient for forecasting and scenario analysis.

To demonstrate the effectiveness of our method, we use a susceptible-infectious-recovered (SIR) ABM as an illustrative example. The proposed method, which we term DeepIMC (Deep Inverse Mapping Calibration), is compared with Approximate Bayesian Computation (ABC), a widely adopted yet computationally demanding calibration technique. The core of the method relies on a supervised learning implementation of the Bidirectional Long Short-Term Memory (BiLSTM) neural network with a mechanistic epidemiological soft constraint. Using DeepIMC, we present a comprehensive simulation study comparing both accuracy and efficiency between these methods.

Furthermore, we introduce \textbf{epiworldRCalibrate}, a software package implemented in R that encapsulates our calibration approach and provides an accessible tool for researchers and practitioners. Importantly, our objective is not to replace Bayesian calibration when full posterior inference is required, but to provide a fast and scalable alternative when the primary goal is accurate epidemic trajectory reconstruction under tight computational constraints. This paper proceeds by first presenting an overview of calibration methods in ABMs, followed by a detailed description of DeepIMC, an extensive simulation study, and concludes with a discussion of findings and implications for future research.

\section{Calibration in ABMs}

Agent-Based Models (ABMs) are computational models in which individuals, or agents, are represented as unique and autonomous entities that interact with one another and with exogenous environments \parencite{railsbackAgentbasedIndividualbasedModeling2019}. ABMs are powerful tools for exploring complex systems and can illuminate subjects such as regulation, investment, and banking activities where agents pursue specific goals. The three fundamental components of an ABM are the agents, the topology (or interaction dynamics), and the environment. Agents are typically distinct in their characteristics and act independently, reflecting heterogeneity and autonomy. The topology specifies how agents are connected and interact, shaping the flow of information and influence within the system. Finally, the environment provides the external context in which agents operate. The overall behavior of the system emerges from the cumulative effects of individual agents and their interactions, often revealing dynamics that cannot be understood by analyzing agents in isolation.

In order to refine and validate the ABM to approximate observed data as much as possible, calibration is needed. Calibration involves adjusting various model parameters, such as agent attributes, decision-making rules, and environmental factors, to align the model's behavior with empirical data or expert knowledge. Calibration requires comparing model outputs to observed data to identify the parameter values that best replicate real-world dynamics. There are various strategies for ABM calibration, including Approximate Bayesian Computation\autocite{toniApproximateBayesianComputation2008, gozzi2022preliminary}, simulated minimum distance \autocite{platt2020comparison}, and surrogate models \autocite{lampertiAgentbasedModelCalibration2018, jalalBayCANNStreamliningBayesian2021, zhang2020validation}. 

Approximate Bayesian Computation (ABC) is a widely used likelihood-free calibration strategy that infers parameter values by repeatedly simulating the model and retaining parameter draws that produce outputs close to the observed data under a chosen discrepancy metric. This makes ABC especially useful when the likelihood is unavailable or computationally intractable. However, because it relies on large numbers of forward simulations, ABC can become prohibitively expensive for agent-based epidemic models, particularly when calibration must be repeated across locations, intervention scenarios, or forecasting horizons.

To address this computational burden, an alternative line of work uses machine learning to construct surrogate models that approximate the mapping from model parameters to simulated epidemic trajectories. Once trained, these models can accelerate calibration by replacing expensive simulations with fast predictions. In contrast, the approach proposed in this paper learns the inverse mapping—from observed epidemic trajectories to the underlying model parameters, which allows rapid recovery of parameter sets that reproduce observed dynamics.

Despite advances in calibration techniques, some limitations persist, including the need for substantial computational resources and the difficulty of handling high-dimensional and complex models. In other words, while ML applications have accelerated calibration, simulation-based algorithms like Markov Chain Monte Carlo are still needed to infer model parameters. Additionally, capturing the whole uncertainty landscape of parameter estimates and effectively incorporating heterogeneity among agents remain ongoing challenges. In applied epidemic modeling, this speed difference can determine whether calibrated models are usable for real-time forecasting or policy evaluation.

\section{Methods}

\subsection{Problem set up}

The Susceptible-Infected-Recovered (SIR) models are parametrized by the recovery rate $p_{recov}$, transmission rate $p_{tran}$, population size $N$ and contact rate $c_{rate}$, which combined with an initial number of infected cases (prevalence), can generate epidemic curves $\text{epi}_t$, \textit{i.e.}, the total number of infected agents over time. More formally, an SIR model can be seen as a random map $M_{SIR}$ from its parameters to its generated epidemic curves, i.e. 
\begin{equation}\label{eq:SIR_mapping}
    M_{SIR}: \theta \to Y,
\end{equation}

\noindent with the specific mapping $M_{SIR}$ being complex and unknown in reality, especially considering a wide range of $\theta$ values. Our task with this paper is to utilize the data $Y$ generated by ABMs with varying parameters $\theta$, to learn the inverse mapping

\begin{equation}\label{eq:ml_mapping}
    M_{SIR}^{-1}: Y \to \theta
\end{equation} 

\noindent in a supervised manner. A successful deployment of DeepIMC  to learn $M_{SIR}^{-1}$ would allow us to feed it observed disease data in $Y$, letting us (i) recover the SIR fundamental parameters $\hat\theta$ and (ii) utilize them to understand the current epidemiological dynamics, such as predicting future number of cases. Throughout this study, population size and recovery rate are assumed to be known or reliably estimated from external sources, consistent with common practice in applied epidemic modeling. Calibration, therefore, focuses on recovering transmission-related parameters from incidence trajectories. We emphasize that learning $M_{SIR}^{-1}$ is not intended to recover uniquely identifiable “true” parameters. In SIR-type models, multiple parameter combinations can generate epidemiologically indistinguishable incidence trajectories. Our goal is therefore to recover parameter sets that are dynamically consistent with the observed data and suitable for forward simulation.


\subsection{Machine Learning Method: DeepIMC}
 
The \textit{calibration} task, consisting of learning the map $M_{SIR}^{-1}$ in (\ref{eq:ml_mapping}) from epidemic trajectories to underlying parameters, naturally calls for sequence models. \textit{Recurrent neural networks} (RNNs) are well suited to this setting because they maintain a hidden state that propagates information across time, allowing the model to encode the growth-peak-decay phases that characterize ABM-generated epidemic curves. Among RNN variants, we adopt a \textit{Long Short-Term Memory} (LSTM) architecture because its gating mechanism (input, forget, output gates) mitigates vanishing gradients and preserves medium-range temporal dependencies over multiple days, which is essential for distinguishing parameter combinations that produce similar early dynamics but diverge near the peak or in the tail. We further use a \textit{bidirectional} LSTM (BiLSTM) since calibration is performed on complete trajectories rather than streaming data; leveraging both forward- and backward-time context improves parameters identifiability from the full curve shape.
 
We train our specific BiLSTM model, defining its data input with what is typically observed for diseases: the population size, recovery rate, and incidence curve. We tasked it to predict key epidemiological parameters: the transmission probability, contact rate, and the basic reproduction number. According to our problem set up in (\ref{eq:ml_mapping}), this formally means:
\begin{equation}
    Y := \{N, p_{recov}, \text{epi}_t\} \qquad \text{and} \qquad \theta := \{ p_{tran}, c_{rate}, R_0\},
\end{equation}
where $\theta$ contains critical parameters for understanding and controlling epidemic outbreaks.
 
\noindent \textbf{Alternative models}. We considered simpler recurrent cells (\textit{e.g.}, vanilla RNNs or GRUs) and more complex attention-based encoders (Transformers). While GRUs are lighter, LSTMs offer a more expressive gating scheme that proved advantageous for our 30-60 day horizons. Transformers are state-of-the-art for long sequences and large training corpora, but in our application--short sequences, a low-dimensional target (3 parameters), and a modest data regime--they introduce substantial parameterization and training overhead without commensurate gains. BiLSTM therefore provides a parsimonious and effective compromise: it captures the relevant temporal structure of ABM incidence curves while remaining stable to train and straightforward to deploy.
 
We also explored Convolutional Neural Networks (CNNs) for our calibration task. While CNNs are typically used for images, we adapted them to time series by using 1D temporal convolutions and global pooling to map incidence trajectories directly to epidemiological parameters. However, our comparative analysis showed that BiLSTMs performed better on this sequential task because they are expressly designed to capture temporal dependencies across the growth--peak--decay phases of an epidemic curve. Beyond predictive performance, LSTM-based models also offered a practical advantage for our data pipeline: they handle variable-length trajectories cleanly via padding together with masking (or packed sequences), so padded time steps are ignored during training. This allowed us to ingest contagion curves of differing lengths without truncation or ad hoc resampling, and thus we proceeded with BiLSTM networks for our calibration approach.
 
\noindent \textbf{Input data}. One of the great benefits of this approach is that the training data can be generated by the Agent-Based Model. Since the goal is to recover the model parameters from the realized datasets within the model, we rely entirely on data simulated from the ABM. Thus, the input to our model consists of daily incidence counts spanning sequences of typically $60$ days, complemented by crucial known time-invariant covariates, namely population size and recovery rate ($p_{recov}$). For most diseases, $p_{recov}$ is typically known, or could be reasonably estimated, although its inclusion in our model only improves $R_0$ prediction marginally, leaving unchanged $p_{tran}$ and $c_{rate}$ prediction accuracy. Our input covariates are normalized using MinMax scaling, facilitating machine model learning by ensuring inputs are on a comparable scale, and the same scalers were applied at test time. Details regarding the specific SIR data-generating process we utilized to train our BiLSTM are described in \autoref{sec:sim-parameters}.
 
\noindent\textbf{Model architecture}. We use three stacked BiLSTM layers with 160 hidden units each, a depth--width combination selected via hyperparameter tuning; making the network deeper or wider did not yield material gains. Bidirectionality allows the model to exploit both past-to-future and future-to-past context across the full epidemic trajectory, improving identifiability of parameters from curve shape. Because this capacity is relatively high compared with our 30--60 day sequences, we regularize the inter-layer representations with dropout applied \emph{between} LSTM layers, which discourages co-adaptation while preserving the cell's within-time memory dynamics. Full details of the architecture and tuning procedure are provided in \autoref{app:training}.
 
\begin{figure}[ht]
\centering
\includegraphics[width=1.06\textwidth]{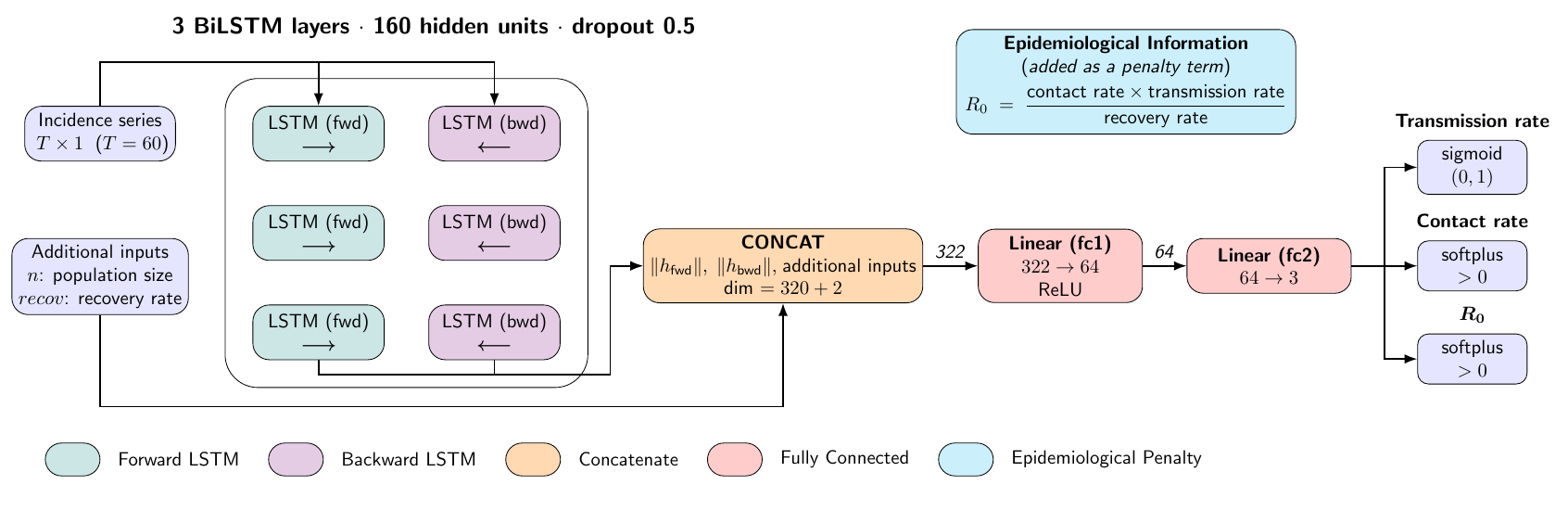}
\caption{Architecture of the proposed DeepIMC architecture.
A univariate incidence time series ($T=60$) is processed through three stacked bidirectional LSTM layers (160 hidden units per direction, dropout rate 0.5).
The final forward and backward hidden representations are concatenated and combined with additional epidemiological inputs (population size $n$ and recovery rate).
The resulting feature vector (dimension $320 + 2 = 322$) is passed through two fully connected layers ($322 \rightarrow 64 \rightarrow 3$), with ReLU activation in the first layer.
The network outputs three epidemiological quantities: transmission rate (sigmoid activation), contact rate (softplus activation), and basic reproduction number $R_0$ (softplus activation).
To enforce epidemiological consistency, the constraint
$R_0 = \frac{\text{contact rate} \times \text{transmission rate}}{\text{recovery rate}}$
is incorporated as a penalty term during training.}
\label{fig:bilstm_diagram}
\end{figure}
 
\noindent \textbf{Output activation function}. The final hidden states from the forward and backward directions of the BiLSTM layers are concatenated with the normalized time-invariant features, creating a rich and informative feature representation. This combined feature set is then passed through two fully connected dense layers: the first dense layer with 64 units employing a ReLU activation function to introduce non-linearity, followed by a final output layer designed to produce epidemiologically interpretable predictions.
 
Output layer activations are carefully selected to enforce biologically meaningful constraints: a sigmoid activation function bounds the transmission probability between 0 and 1, and softplus activation functions ensure positivity for the predicted contact rate and the basic reproduction number ($R_0$).
 
\noindent \textbf{Training regime and loss function}. The model is trained with a composite loss function $\mathcal{L}$ comprising Mean Squared Error (MSE) and a penalty term designed to maintain epidemiological consistency. Specifically, the penalty encourages the predicted parameters to satisfy the theoretical relationship $R_0 \times \text{recovery rate} \approx \text{transmission probability} \times \text{contact rate}$. The loss function can be mathematically represented as:
\begin{flalign}\label{eq:loss-pop}
\mathcal{L}(\phi)
= \mathbb{E}\Big[
  \underbrace{\|\hat\theta_{\phi}(Y)-\theta\|_2^2}_{\text{MSE term}}
  \;+\;
  \lambda\,\underbrace{\big(\hat R_{0,\phi}(Y)\,\gamma \;-\; \hat p_{\text{tran},\phi}(Y)\,\hat c_{\text{rate},\phi}(Y)\big)^2}_{\text{epidemiological consistency penalty}}
\Big],
\end{flalign}
where $\theta = \{ p_{tran}, c_{rate}, R_0\}$, $Y = \{N, p_{recov}, \text{epi}_t \}$, and $\phi$ are BiLSTM parameters. Full details of the optimization procedure and hyperparameter selection, including the chosen value of $\lambda$, are provided in \autoref{app:training}.
 
Overall, this carefully designed approach not only ensures high predictive accuracy and computational efficiency but also guarantees epidemiological coherence in parameter estimation, making our model particularly suitable for real-time calibration and informing public health decision-making.
 
Furthermore, the implementation of this calibration method is publicly available on GitHub at \url{https://github.com/sima-njf/epiworldRcalibrate}.
 
\section{Simulation Study}
Our simulation study evaluates and benchmarks the performance of Approximate Bayesian Computation (ABC), particularly, the likelihood-free Markov Chain Monte Carlo (LFMCMC) algorithm\autocite{marjoramMarkovChainMonte2003}, as implemented in the R package \textbf{epiworldR} package\autocite{meyerEpiworldRFastAgentBased2023} against a data-driven DeepIMC estimator for parameter estimation and epidemic forecasting within the agent-based SIR modeling framework. Specifically, we executed a comprehensive set of simulations involving 5,000 independent parameter scenarios. For each scenario, we generated synthetic epidemic trajectories and subsequently assessed each method's ability to recover underlying epidemiological parameters, accurately predict infection dynamics, and efficiently calibrate models. Implementation details of the ABC procedure, including proposal mechanisms and discrepancy measures, are provided in Appendix~\ref{sec:abc-appendix}.
The parameter values used to generate the testing datasets were the same as those we used for training the BiLSTM (see \autoref{sec:sim-parameters}). It is important to highlight two points: First, we use a new draw from the model parameters to generate the testing dataset, and second, the prior distributions used to generate the training and testing sets are sufficiently wide to create highly heterogeneous realized epidemiological curves. Since the ABM is a stochastic model, no two curves with the same parameter set are equal.
The testing procedure consisted of comparing how well each method was at recovering (a) the parameter values used to generate the incidence curve and (b) the curve itself. For each one of the 5,000 drawn parameters, we did the following:
\begin{enumerate}
  \item Simulate a 60‑day epidemic of daily infected counts using the agent‑based SIRCONN model--perfect mixing ABM--in the epiworldR package.
  \item ABC and our approach were then used to calibrate the model to the generated infected counts (the daily incidence). Calibration using our proposed method was done using the \textbf{epiworldRCalibrate} package, and ABC calibration was done using the implementation available in \textbf{epiworldR}.
\end{enumerate}
After running the calibration procedure, we compared both parameter estimates and goodness of fit of the epidemiological curves for the concurrent models in the following ways:\\
\noindent\textbf{Parameter recovery}. For each method, we calculated bias, relative bias, and Mean Absolute Error of each parameter. Notwithstanding, parameter recovery is crucial for an inferential process. It is essential for the reader to note that our model's primary objective is to reproduce the observed epidemiological curve. Because the contact rate and transmission rate interact in the ABM, it is not possible to exactly recover both. We further this point later in the Discussion section of the paper.\\
\noindent\textbf{Predictive bias.} After estimating parameters with the two approaches, we forward simulate 100 new curves using the calibrated parameters and compare the predicted trajectories against those from the ground-truth parameters. This allows us to assess the downstream predictive performance of the calibrated parameters in realistic settings:  \(\Delta_t= I_t - \widehat I_t\) and \(\Delta_t/I_t\). In addition to bias and relative bias, we also computed the empirical 95\% credible-interval coverage, as well as computation time (wall-clock seconds per ABC and DeepIMC).
All the code used to conduct the experiment is available in the following GitHub repository: \url{https://github.com/UofUEpiBio/Calibrate_ABM}.
\section{Results: Comparing with ABM}

\subsection{Predictive Evaluation: ABC vs DeepIMC}

We systematically assess the predictive accuracy of two distinct modeling approaches, Approximate Bayesian Computation and DeepIMC, in forecasting infectious disease dynamics under scenarios characterized by parameter uncertainty. Each approach was rigorously tested using 5,000 distinct parameter sets, with each set evaluated through 100 stochastic 60-day simulations, ensuring comprehensive coverage of diverse epidemiological conditions and parameter configurations.

Figure~\ref{fig:coverage} below illustrates comparative predictive trajectories generated by ABC and DeepIMC models for infected individuals, cumulative infected, and susceptible populations. These figures include clearly marked 95\% prediction intervals, highlighting variability and predictive confidence. Notably, the DeepIMC model consistently produces tighter and more precise credible intervals, signifying a superior capability to accurately forecast disease progression dynamics over the entire simulation horizon.

\begin{figure}
\centering
\includegraphics[width=1\textwidth]{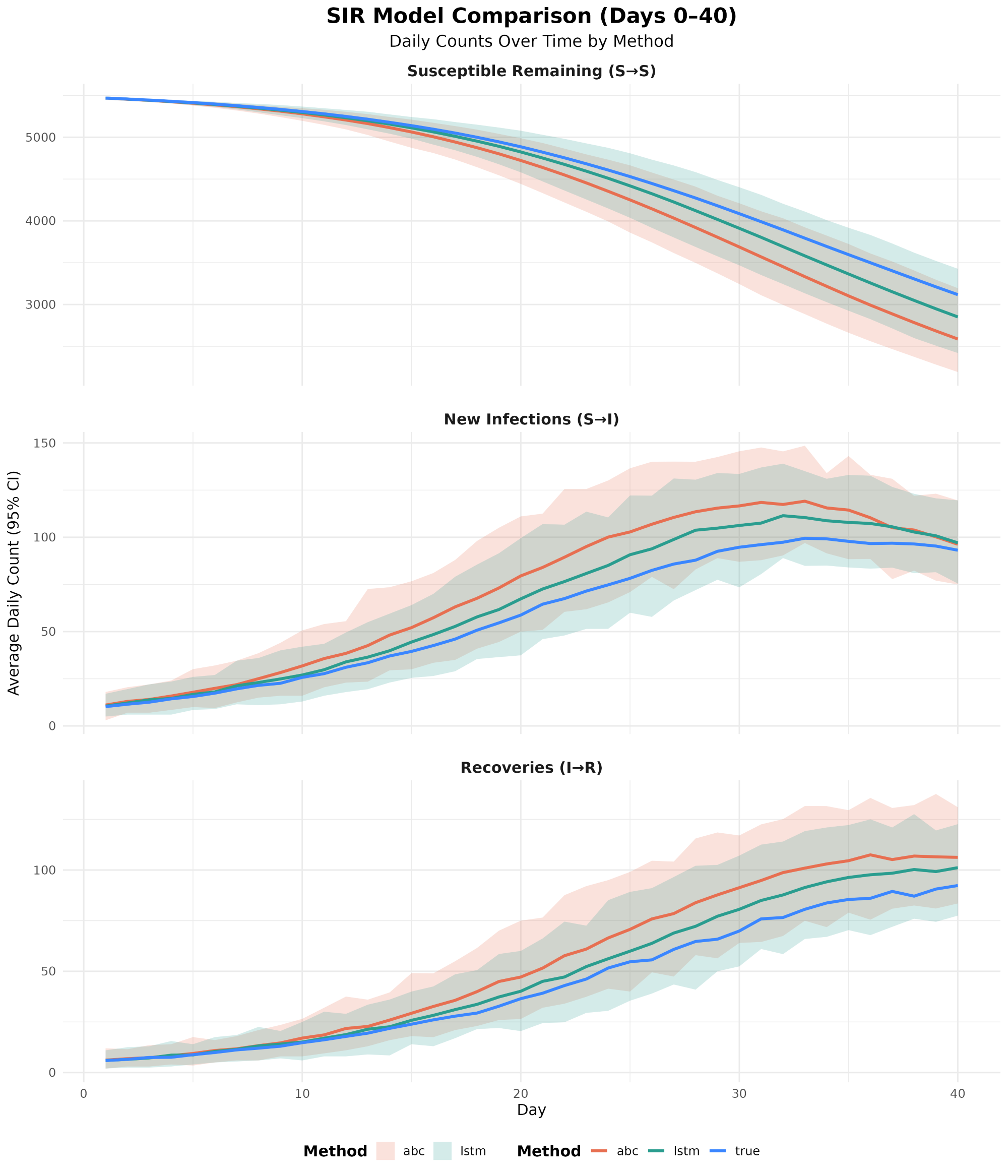}
\caption{Predicted epidemic curves showing susceptible, infected, and recovered populations over 40 days for a representative subset of parameter simulations. Credible intervals indicate model uncertainty.}
\label{fig:epidemic_curves}
\end{figure}

These findings demonstrate that the DeepIMC model is not only more efficient but also better calibrated, offering a more reliable predictive envelope for public health decision-making. The visual and numerical evidence points to DeepIMC's advantage in uncertainty quantification, which is crucial for scenario planning in rapidly evolving epidemiological contexts.
These plots provide further insight into the differences in predictive uncertainty between ABC and DeepIMC.

Figure~\ref{fig:coverage} compares the bias and coverage performance of the ABC and DeepIMC methods for epidemiological parameter estimation using simulated infected count data.
The top-left panel shows the mean percentage bias over the 60-day period, with DeepIMC maintaining near-zero bias while ABC exhibits substantial negative bias, particularly between days 10 and 20. The top-right panel shows coverage over time, with both methods remaining close to the nominal 95\% coverage level. The bottom-left panel summarizes key statistics, indicating that DeepIMC achieves lower mean absolute bias and RMSE compared to ABC. The bottom-right panel compares overall coverage, showing both methods achieve values close to the 95\% target, though DeepIMC provides consistently more accurate predictions with reduced bias.

\begin{figure}
\centering
\includegraphics[width=1\textwidth]{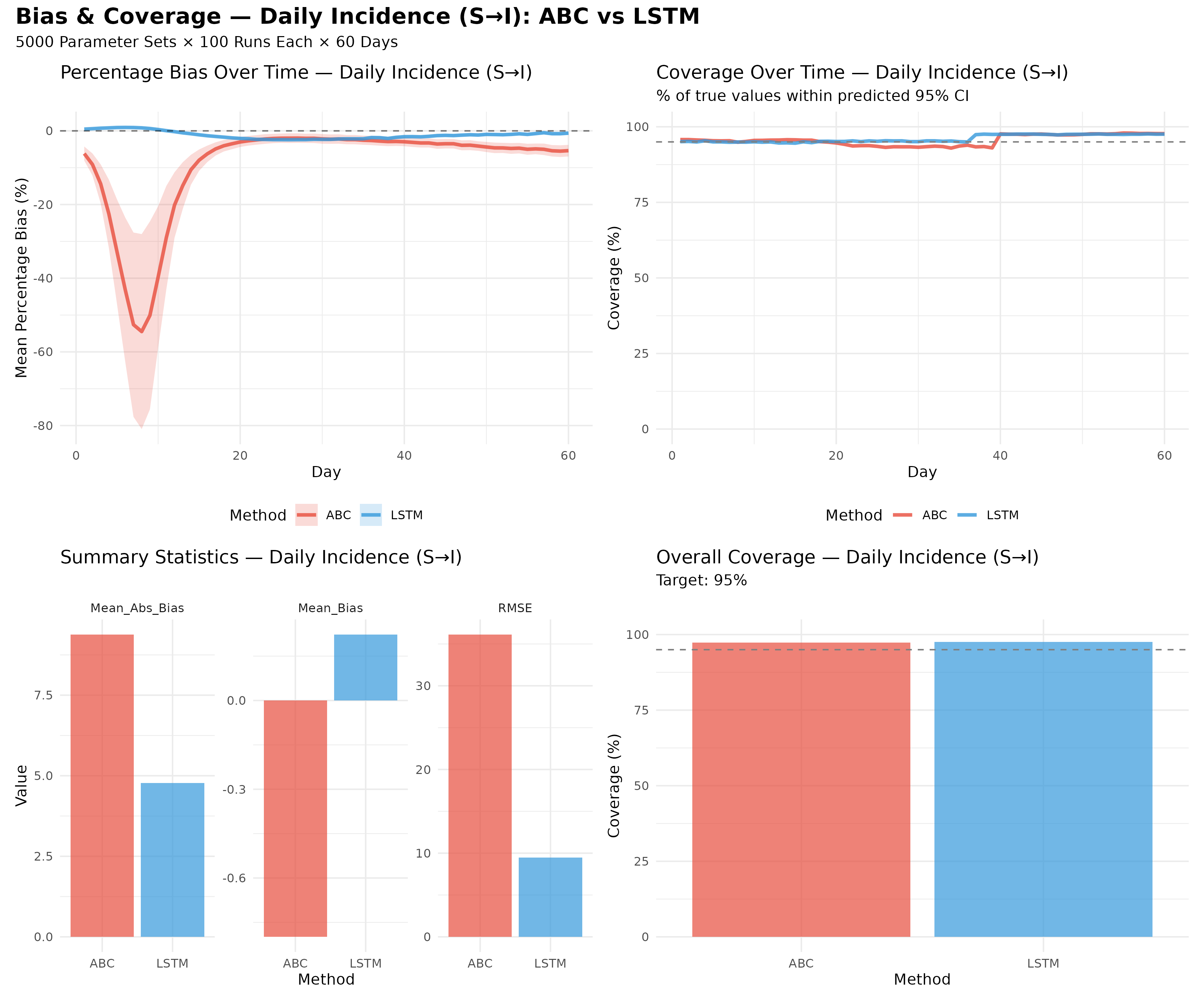}
\caption{Comparative analysis of predictive bias and coverage between ABC and DeepIMC methods over a 60-day forecast period. The DeepIMC model consistently demonstrates lower bias and superior coverage accuracy.}
\label{fig:coverage}
\end{figure}

\begin{figure}[!ht]
\centering
\includegraphics[width=0.9\textwidth]{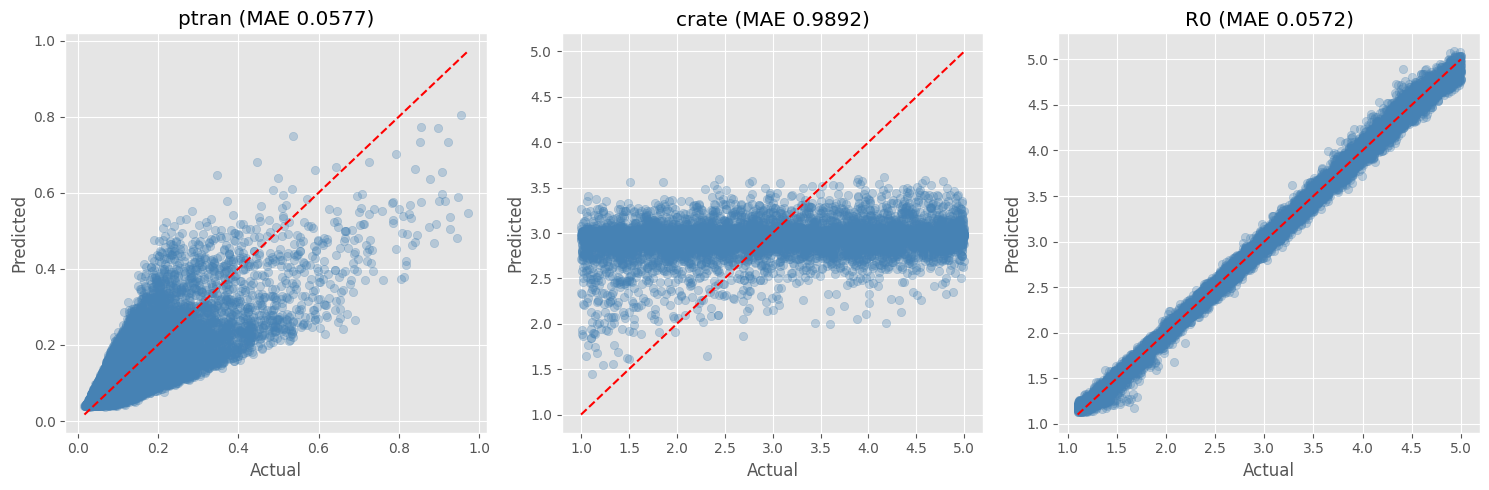}
\caption{Performance of the DeepIMC model trained on ABM-generated epidemiological time-series data in predicting transmission probability ($p_{\text{tran}}$), contact rate ($c_{\text{rate}}$), and basic reproduction number ($R_0$). Scatter plots compare predicted versus actual values for each parameter, with the red dashed line indicating the ideal 1:1 correspondence. The Mean Absolute Errors (MAEs) are reported in parentheses for each metric.}
\label{fig:network_topology}
\end{figure}

Figure~\ref{fig:network_topology} illustrates the performance of the DeepIMC model in predicting transmission probability ($p_{\text{tran}}$), contact rate ($c_{\text{rate}}$), and basic reproduction number ($R_0$) from ABM-generated epidemiological time-series data. Predicted values are plotted against actual values, with the red dashed line indicating the ideal 1:1 correspondence. The model achieves low prediction errors for $p_{\text{tran}}$ (MAE = 0.0577) and $R_0$ (MAE = 0.0572), demonstrating strong agreement with the true values. However, predictions for $c_{\text{rate}}$ show substantial bias and reduced accuracy (MAE = 0.9892), indicating challenges in learning contact rate dynamics from the available input features.

\subsubsection{Parameter Recovery Accuracy}

Accurate parameter estimation is critical in epidemiological modeling, influencing the robustness and utility of predictive models. We evaluated the ability of ABC and DeepIMC methods to recover key epidemiological parameters by analyzing their estimation errors. Table~\ref{tab:parameter_recovery} provides a detailed summary of the performance metrics.

\begin{table}
\centering
\caption{Detailed summary of estimation error metrics for epidemiological parameters by ABC and DeepIMC methods.}
\label{tab:parameter_recovery}
\begin{tabular}{llrrrr}
\\
\toprule
\textbf{Parameter} & \textbf{Method} & \textbf{MAE} & \textbf{RMSE} & \textbf{Mean Bias} & \textbf{Median Bias} \\
\midrule
$R_0$             & ABC     & 1.8900  & 2.8400  &  1.7600   &  0.0075   \\
                  & DeepIMC & 0.0592  & 0.0780  &  0.0145   &  0.0123   \\
\midrule
Contact rate      & ABC     & 6.2000  & 11.6000 &  5.7500   &  2.9500   \\
                  & DeepIMC & 1.0400  & 1.2400  & $-0.5180$ & $-0.5140$ \\
\midrule
Transmission rate & ABC     & 0.1290  & 0.1870  & $-0.1010$ & $-0.0730$ \\
                  & DeepIMC & 0.0719  & 0.0979  &  0.0035   &  0.0249   \\
\bottomrule
\end{tabular}
\end{table}

\begin{figure}[!ht]
\centering
\includegraphics[width=1.05\textwidth]{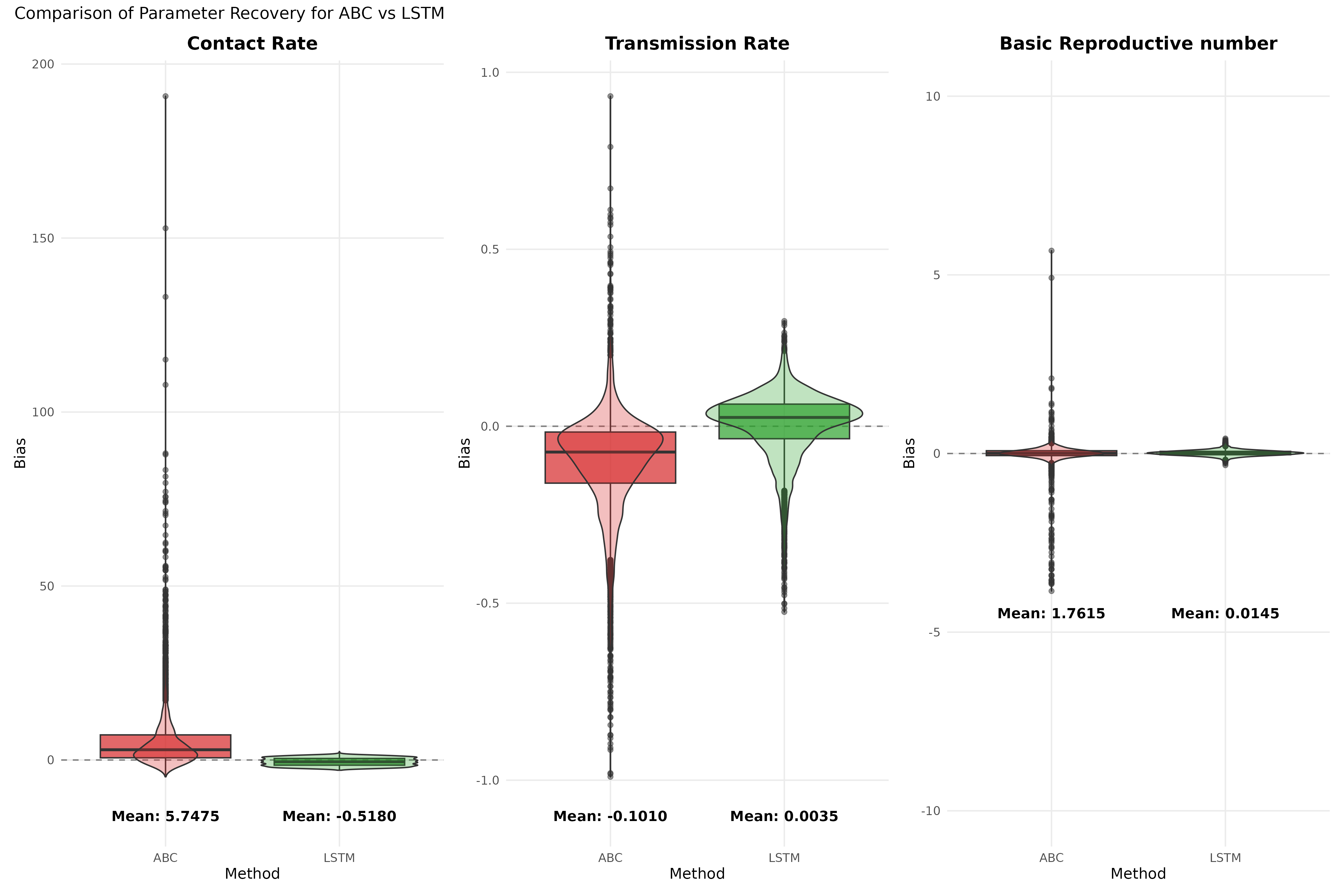}
\caption{Violin and box plots showing the distribution of parameter estimation bias for ABC and DeepIMC methods across three key epidemiological parameters: Contact Rate, Transmission Rate, and Basic Reproduction Number ($R_0$).}
\label{fig:param_recovery_violin}
\end{figure}

Figure~\ref{fig:param_recovery_violin} provides a visual comparison of the bias distribution for each parameter estimated using ABC and DeepIMC methods. The plots combine violin and boxplot representations, offering both distribution shape and summary statistics such as median and quartiles.

Across all three parameters, DeepIMC demonstrates a consistently narrower spread in the distribution of bias, and its median bias is closer to zero. This reflects both improved central accuracy and lower variability. For instance, in the case of the contact rate, ABC exhibits a long right tail---indicating the tendency to substantially overestimate in certain scenarios---whereas DeepIMC remains tightly centered with less skew. Similarly, for transmission probability and $R_0$, the DeepIMC model again shows a tighter concentration of estimates around the true values.

Regarding calibration speed, Table~\ref{tab:comp_time} presents the average wall-clock time required to perform a single model calibration run for both the Approximate Bayesian Computation and the DeepIMC methods. As expected, ABC takes significantly longer to finalize the calibration process, about thirty times longer than our ML-based method. Although we are indicating here that it takes about 77 seconds to calibrate the model using ABC, readers should consider that we are testing this in a very simple model, a perfect mixing SIR. In other scenarios, where a single run of an ABM could take hours, the speedup gains can be significantly larger; the bulk ML-based method's computational complexity lies in the estimation process rather than the prediction algorithm.

\begin{table}
\centering
\caption{Wall-clock time per calibration run for ABC vs.\ DeepIMC methods.}
\label{tab:comp_time}
\begin{tabular}{@{}lr@{}}
\toprule
\textbf{Method} & \textbf{Computation Time} \\
\midrule
Approximate Bayesian Computation (ABC) & 77.40\,seconds \\
DeepIMC                                &  2.35\,seconds \\
\bottomrule
\end{tabular}
\end{table}

\noindent Collectively, these results underline the substantial advantages of the DeepIMC method in predictive accuracy, parameter recovery, and computational efficiency compared to traditional ABC methods in modeling infectious disease dynamics. The tight predictive intervals, improved estimation accuracy, and computational scalability make DeepIMC an especially compelling option for real-time applications, such as outbreak forecasting and public health intervention planning.
\subsection{Application to Real COVID-19 Incidence Data}
To complement the simulation-based evaluation, we apply both calibration approaches to observed COVID-19 incidence data from the state of Utah. The data consist of daily confirmed new reported COVID-19 cases aggregated at the state level and were obtained from the Utah Department of Health public COVID-19 surveillance repository. These counts reflect reported laboratory-confirmed cases by report date and serve as a commonly used proxy for underlying infection dynamics in real-time epidemic analysis. We focus on a contiguous 60-day period spanning early March through early May 2020, corresponding to the initial epidemic growth and early mitigation phase in Utah.

Daily incidence counts are treated as observations of new infections and are used as the sole calibration target for both ABC and the proposed DeepIMC method. For both approaches, calibrated parameter estimates are propagated forward through stochastic simulations of an agent-based Susceptible–Infected–Recovered (SIR) model to generate predictive trajectories and empirical 95\% prediction intervals. While this SIR model is intentionally simple and does not account for behavioral changes, spatial heterogeneity, reporting delays, or policy interventions, it provides a transparent and controlled setting for comparing calibration behavior, uncertainty propagation, and computational efficiency across methods.

\begin{figure}[!ht]
\centering
\includegraphics[width=1\textwidth]{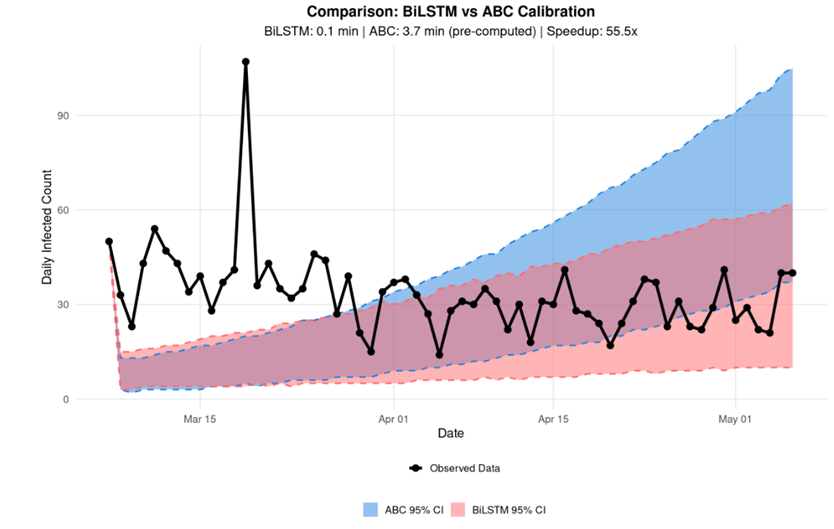}
\caption{Application of ABC and DeepIMC calibration to observed COVID-19 incidence data from Utah. The black line shows daily reported new cases over a 60-day period from early March to early May 2020. Shaded regions denote empirical 95\% prediction intervals obtained from 2,000 stochastic forward simulations of an agent-based SIR model using parameters calibrated via Approximate Bayesian Computation (ABC) and the DeepIMC inverse mapping. While ABC yields wider uncertainty bands reflecting posterior variability, the DeepIMC approach produces tighter predictive intervals with comparable coverage at substantially lower computational cost.}
\label{fig:utah_abc_bilstm}
\end{figure}

Figure~\ref{fig:utah_abc_bilstm} compares the resulting predictive intervals against the observed Utah incidence data. The ABC-based calibration produces relatively wide prediction bands, reflecting posterior uncertainty propagated through repeated stochastic simulation. In contrast, the DeepIMC-calibrated model yields substantially tighter prediction intervals while maintaining comparable empirical coverage. Importantly, these results are obtained at a fraction of the computational cost: once trained, DeepIMC produces calibrated parameter estimates in seconds via a single forward pass, whereas ABC requires repeated simulation and careful tuning of discrepancy measures, tolerance thresholds, and proposal mechanisms even in this low-dimensional setting. Both methods show some discrepancy relative to the observed early-phase trajectory, which is expected given that the early COVID-19 epidemic in Utah was shaped by rapidly shifting reporting practices, behavioral responses, and intervention timing that a homogeneous SIR model cannot capture. This application highlights the practical advantages of machine learning–based inverse calibration for real-time epidemic monitoring, rapid forecasting updates, and repeated calibration across locations or scenarios where computational constraints are binding.

\newpage
\section{Discussion}

In this study, we introduced and evaluated an ML-based calibration framework that improves accuracy and dramatically accelerates the process. In our application, we used DeepIMC --- a bidirectional Long Short-Term Memory (BiLSTM) neural network model --- to achieve fast, reliable, and scalable parameter estimation. Our comprehensive simulation study, encompassing 5,000 simulated epidemic scenarios, clearly demonstrates the strengths of the DeepIMC approach compared to traditional ABC calibration methods.

A critical observation emerging from our analysis is the inherent challenge in simultaneously identifying the three parameters---contact rate, recovery rate, and transmission rate---with high precision, even with sophisticated machine learning architectures and increased data volume. This difficulty arises because at least one parameter tends to default to its average estimate, suggesting a fundamental identifiability constraint within the SIR model framework. Notably, this limitation does not significantly affect the practical utility of our predictions, as the calibrated parameters still produce accurate epidemic curves. This outcome implies that the underlying dynamics of incidence curves generated by SIR models may effectively possess only two degrees of freedom, reinforcing the notion that accurate incidence predictions can be made without precisely estimating every individual parameter. Although in our framework the parameters $\theta$ (contact rate, recovery rate, transmission rate) function primarily as nuisance parameters--estimated only to generate reliable epidemic curves--it is still possible to quantify uncertainty in their values. Future work could extend our calibration framework to incorporate statistical inference on $\theta$, for instance, by generating distributions of predicted parameter sets rather than single point estimates. One practical approach would be to train multiple calibrators on bootstrapped datasets (e.g., 100 models), thereby producing a collection of plausible parameter estimates. This ensemble-based strategy would enable the construction of credible intervals for $\theta$, providing additional insight into parameter uncertainty while preserving the predictive focus of our modeling.

Our evaluation revealed several distinct advantages of the DeepIMC framework.\\ First, DeepIMC \textbf{substantially improved parameter recovery accuracy compared to ABC-only methods}. Specifically, DeepIMC resulted in lower mean absolute error and root-mean-square error across all targeted parameters, with error distributions consistently centered around zero and exhibiting minimal dispersion. In contrast, ABC methods demonstrated broader error distributions, heavier tails, and systematic biases, particularly evident in parameters such as contact rate and $R_0$.

Second, the \textbf{predictive performance of DeepIMC was superior}. Epidemic trajectories derived from DeepIMC-calibrated models consistently showed reduced bias and narrower 95\% prediction intervals, while maintaining coverage rates near the nominal 95\%. This indicates that DeepIMC achieves more precise yet reliable forecasting capability compared to ABC.

Third, \textbf{computational efficiency represented a notable strength of DeepIMC}. DeepIMC-based calibration completed within seconds on standard processors without parallel computing requirements, in sharp contrast to ABC--LFMCMC methods that necessitated over 77 seconds. This dramatic reduction in computational time significantly enhances the practical applicability of our approach, especially in real-time epidemic forecasting scenarios or situations requiring simultaneous evaluations across multiple settings.

One of the major benefits of employing a neural network for inverse modeling is its \textbf{single-training requirement}. After initial training, DeepIMC can rapidly produce parameter estimates and associated uncertainty intervals through a simple forward pass. Unlike ABC methods, which demand extensive simulation for each parameter exploration, DeepIMC effectively exploits learned patterns from historical simulations. Additionally, neural networks excel at capturing complex nonlinear interactions among parameters, an essential feature for accurately modeling disease dynamics that often challenge traditional statistical sampling methods.

Looking forward, extending DeepIMC to accommodate real-world datasets, which frequently exhibit variable-length or streaming infection trajectories, is a promising direction. Enhancing the BiLSTM architecture with masking layers or attention mechanisms could enable the model to selectively process observed data segments, thereby improving robustness and flexibility in real-world surveillance and response contexts.

\section{Conclusions}
Agent-based models (ABMs) have become indispensable tools across diverse scientific fields, particularly in public health, where they enable detailed scenario modeling of disease outbreaks. Despite their widespread application, calibrating ABMs remains computationally intensive, necessitating the development of methods to expedite the process and improve accuracy. Existing approaches typically rely on surrogate modeling to approximate the ABM, thereby creating a direct mapping from model parameters to observed outcomes.

In this paper, we introduced DeepIMC, an innovative method that reverses this paradigm, constructing an inverse mapping directly from observed data to model parameters. We implemented DeepIMC using a Bidirectional Long Short-Term Memory (BiLSTM) neural network, trained exclusively on simulated data. This allowed us to develop a generalized learner capable of accurately calibrating an SIR-type ABM across various epidemiological scenarios.

Through a comprehensive simulation study, we benchmarked DeepIMC against Approximate Bayesian Computation (ABC), a widely used but computationally demanding calibration technique. Results indicated that DeepIMC consistently outperformed ABC, achieving superior accuracy in recovering true model parameters and epidemic curves, while significantly reducing computational demands. Our methodology has been encapsulated in the \texttt{epiworldRCalibrate} R package, making it readily accessible for practical use.

Nevertheless, significant challenges remain. Issues such as parameter identifiability and the extension of our approach to more complex models, including mixing models and SEIR frameworks, require further investigation. Addressing these challenges presents promising avenues for future research, potentially extending the applicability and impact of ABM calibration methods in public health and beyond.

In conclusion, DeepIMC represents a substantial advancement in ABM calibration, offering notable improvements in accuracy, efficiency, and computational practicality. Our results strongly advocate for its adoption in epidemiological modeling and potentially other domains reliant on simulation-based model calibration and parameter estimation. By learning the inverse mapping from epidemic trajectories to model parameters, DeepIMC provides a fast and scalable alternative to traditional simulation-based calibration, facilitating real-time, large-scale applications and informed decision-making in complex modeling scenarios.

\textcolor{blue}{
\section{Code Availability}
All code required to reproduce the simulation study and results is publicly available at \url{https://github.com/sima-njf/epiworldRcalibrate} and \url{https://github.com/UofUEpiBio/Calibrate_ABM}.
}

\section{Acknowledgments}
This work was made possible by cooperative agreement CDC-RFA-FT-23-0069 from the CDC's Center for Forecasting and Outbreak Analytics. Its contents are solely the responsibility of the authors and do not necessarily represent the official views of the Centers for Disease Control and Prevention.

\printbibliography

\appendix

\section{Simulation of Parameters and Data}\label{sec:sim-parameters}

\subsection{Parameter Setup}

All parameters are randomly drawn for each simulation as follows:
\[
\begin{aligned}
N\ (\text{population size}) &\sim \mathrm{Uniform}(5000,\ 10000), \\
p_{recov} \, (\text{recovery rate}) &\sim \mathrm{Uniform}(0.071,\ 0.25), \\
\text{Prevalence} &\sim \frac{\mathrm{Uniform}(100,\ 2000)}{N}, \\
R_0  &\sim \mathrm{Uniform}(1,\ 5), \\
c_{rate} \, (\text{contact rate}) &\sim \mathrm{Uniform}(5,\ 15), \\
p_{tran} \, (\text{transmission rate}) &= \frac{R_0 \cdot p_{recov}}{c_{rate}} \quad (0 \leq p_{tran} \leq 1).
\end{aligned}
\]

Table~\ref{tab:params_setup} summarizes these distributions.

\begin{table}[ht]
\centering
\caption{Parameter setup for training scenarios.}
\label{tab:params_setup}
\begin{tabular}{llm{.3\linewidth}<\raggedleft}
\toprule
Parameter & Description & Distribution / Value \\
\midrule
\(N\) & Population size (known) & {Uniform}(5000,\ 10000) \\
Recovery rate & Recovery rate (known) & {Uniform}(0.071,\ 0.25) \\
Prevalence & Initial infected proportion & Uniform(100,2000)/N \\
Contact rate & Effective contacts per person & Uniform(5,15) \\
\(R_0\) & Basic reproduction number & Uniform(1,5) \\
Transmission rate & Transmission probability per contact & \(R_0\times p_{recov} / c_{rate}\) \\
\bottomrule
\end{tabular}
\end{table}

\subsection{Preparation of Incidence Data}
Each simulated scenario yields a 60‐day infected‐count trajectory.  We transform it into a \(1\times60\) array:
\begin{itemize}
  \item  daily incidence (new infections).
\end{itemize}
Table~\ref{tab:dataarray} defines these rows.

\begin{table}[ht]
\centering
\caption{Structure of the data array for each scenario.}
\label{tab:dataarray}
\begin{tabular}{ll}
\toprule
Row & Description \\
\midrule
1 & Daily incidence (new cases each day) \\
\bottomrule
\end{tabular}
\end{table}

\section{ABC details}\label{sec:abc-appendix}

The ABC procedure focuses on estimating three key parameters: contact rate, recovery rate, and transmission rate, while keeping prevalence and $R_0$ fixed during calibration. Since the model we are simulating is fairly simple, here we use the entire simulation outcome as the target statistic.

Each LFMCMC chain runs for 2,000 iterations. The first 1,000 are discarded as burn-in, and the posterior median of the accepted samples is used as the point estimate. 

\begin{itemize}
  \item \textbf{Observed data:} For each simulation, we generate a 60-day trajectory of infected counts using true parameters. This serves as the reference data for calibration.

  \item \textbf{Forward simulator:} We define a function \texttt{simulate\_epidemic\_calib()} that simulates a 60-day epidemic given a set of candidate parameters and returns the daily counts of infected individuals.

  \item \textbf{Summary statistic:} The full time series (identity function) is used directly, retaining maximum information without dimensionality reduction.

  \item \textbf{Parameter proposal mechanism:} At each iteration, new candidate parameters are proposed using the following transformations:
  \[
  \begin{aligned}
    c_{rate}' &\leftarrow c_{rate} \cdot e^{\mathcal{N}(0, 0.1^2)}, \\
    p_{recov}' &\leftarrow p_{recov} \cdot e^{\mathcal{N}(0, 0.1^2)}, \\
    p_{trans}' &\leftarrow \text{logit}^{-1}\left(\text{logit}(p_{trans}) + \mathcal{N}(0, 0.1^2)\right).
  \end{aligned}
  \]

  \item \textbf{Discrepancy measure:} The similarity between simulated and observed trajectories is measured using Euclidean distance. The kernel weight assigned to a proposed parameter set is:
  \[
  K_\varepsilon = \exp\left(-\frac{\|S' - S_{\rm obs}\|_2^2}{2\varepsilon^2}\right), \quad
  \varepsilon = 0.05 \cdot \|S_{\rm obs}\|_2,
  \]
  where $S'$ is the simulated series and $S_{\rm obs}$ is the observed one.
\end{itemize}
\section{Training Details}\label{app:training}
 
The Adam optimizer was used with a learning rate of approximately $2.77 \times 10^{-4}$. All hyperparameters  including the learning rate, number of BiLSTM layers (3), hidden units per layer (160), and dropout rate (0.5) were selected via Optuna-based hyperparameter optimization. The penalty weight $\lambda$ was tuned jointly with the remaining hyperparameters, yielding $\lambda = 1.77 \times 10^{-4}$, reflecting that the epidemiological consistency penalty serves as a light regularizer rather than a dominant training signal. The model was trained with a batch size of 64 for up to 100 epochs, with early stopping based on validation loss to prevent overfitting.
 
\end{document}